\documentclass[conference]{IEEEtran}
\IEEEoverridecommandlockouts
\usepackage{cite}
\usepackage{amsmath,amssymb,amsfonts}
\usepackage{algorithmic}
\usepackage{graphicx}
\usepackage{textcomp}
\usepackage{xcolor}
\usepackage{soul}
\usepackage{float}
\usepackage{lscape}
\usepackage[section]{placeins}
\makeatletter
\AtBeginDocument{%
  \expandafter\renewcommand\expandafter\subsection\expandafter{%
    \expandafter\@fb@secFB\subsection
  }%
}
\makeatother
\def\BibTeX{{\rm B\kern-.05em{\sc i\kern-.025em b}\kern-.08em
    T\kern-.1667em\lower.7ex\hbox{E}\kern-.125emX}}
\begin{document}

\title{Deep Generative Models: Deterministic Prediction with an Application in Inverse Rendering
{\footnotesize \textsuperscript{*}Note: Sub-titles are not captured in 
Xplore and should not be used}
\thanks{Identify applicable funding agency here. If none, delete this.}
}

\author{\IEEEauthorblockN{Shima Kamyab}
\IEEEauthorblockA{\textit{Dept of Comp. Sci., Eng. and IT } \\
\textit{Shiraz University}\\
Shiraz, Iran \\
sh.kamyab@cse.shirazu.ac.ir}
\and
\IEEEauthorblockN{Rasool Sabzi}
\IEEEauthorblockA{\textit{Dept of Comp. Sci., Eng. and IT } \\
\textit{Shiraz University}\\
Shiraz, Iran \\
sabzi@cse.shirazu.ac.ir}
\and
\IEEEauthorblockN{Zohreh Azimifar}
\IEEEauthorblockA{\textit{Dept of Comp. Sci., Eng. and IT } \\
\textit{Shiraz University}\\
Shiraz, Iran\\
azimifar@cse.shirazu.ac.ir}

}

\maketitle

\begin{abstract}

Deep generative models are stochastic neural networks capable of learning the distribution of data so as to generate new samples. Conditional Variatonal Autoencoder (CVAE) is a powerful deep generative model aiming at maximizing the lower bound of training data log-likelihood. In this structure, there is appropriate regularizer, which makes it applicable for suitably constraining the solution space in solving ill-posed problems and providing high generalization power. Considering the stochastic prediction characteristic in CVAE, depending on the problem at hand, it is desirable to be able to control the uncertainty in CVAE predictions. Therefore, in this paper we analyze the impact of CVAE's condition on the diversity of solutions given by our designed CVAE in 3D shape inverse rendering as a prediction problem. The experimental results using Modelnet10 and Shapenet datasets and comparison with several recent methods show the appropriate performance of our designed CVAE and verify the hypothesis: \emph{``The more informative the conditions in terms of object pose are, the less diverse the CVAE predictions are}".
\end{abstract}

\begin{IEEEkeywords}
Generative models for prediction, Deep generative models, 3D shape inverse rendering
\end{IEEEkeywords}

\section{Introduction}
Generative models include a broad domain of machine learning techniques with the aim of using statistical methods to learn the underlying distribution of a set of real data to generate similar synthetic data from the learned distribution. More formally, suppose sample $x$ is obtained from an unknown distribution $P_{gt}(x)$. The objective is to learn the distribution $P$ so that sample $x'$ drawn from $P$ is as close to $x$ as possible.  

\par
The traditional generative models suffer from three main drawbacks \cite{doersch2016tutorial}:
\begin{itemize}
\item In some cases, they need large amount of information about the structure of data.
\item The training method used in these techniques may causes high level of uncertainty and therefore, the resulted synthetic samples may be unfeasible.
\item Most of these methods suffer from high computational complexity.
\end{itemize}

In recent years, after the advent of the field of deep learning, deep generative models, which exhibit strong performance in modeling complex high dimensional distributions of text or image, have attracted a large body of interest in the literature. Due to their powerful nonlinear approximation, these models are appropriate tools for estimating the density of complicated and high dimensional data. Deep generative models are applied in many fields including text generation~\cite{semeniuta2017hybrid, hu2017controllable, karras2017progressive}, latent space learning~\cite{chen2016infogan}, image denoising~\cite{chen2018image}, image in-painting~\cite{yeh2017semantic}, super-resolution~\cite{ledig2017photo}, etc.

Two of the most well-known and efficient deep generative models are Variational AutoEncoders (VAEs) and Generative Adversarial Networks (GANs). The VAE's objective is to maximize the lower bound of training data's likelihood function and GANs aim at achieving an equilibrium between their two adversarial components known as \emph{Generator} and \emph{Discriminator}.\par

In this paper our focus is on VAE models, where the data generation problem can be formulated in the form of a Bayesian model~\cite{kingma2013auto}. The main contribution of VAE is to maximize the likelihood function of training data on the whole generative process by conditioning the output distribution on some latent variable learned by using the information from training data~\cite{doersch2016tutorial}. 
Although VAEs could generate feasible samples, the models have no control on the generated output. Conditional Variational AutoEncoder (CVAE) as a modified VAE addresses this limitation by utilizing some condition information to control the type of the output. In the case of CVAE, all distributions are conditioned on some measurement, called $\underline{m}$, and this condition is used as an input to both encoder and decoder components.

\par
CVAE is a stochastic model capable of learning multi-modal distributions, \textit {i.e.}, finding a one-to-many mapping resulting different solutions for the problem. This characteristic of CVAE is applicable in many problems like structural classification~\cite{sohn2015learning}. On the other hand, for some situations like ill-posed problems, it is desirable to have more deterministic solutions, \textit{e.g.}, 3D shape inverse rendering with complicated solution space, needs the solutions to be deterministic in terms of personal identity so that they could recover the true shape accurately~\cite{taigman2014deepface}. Therefore, if some mechanism could be provided to control the diversity of the solution space in CVAE, it will be more effective in different problem domains. In this paper we analyze the impact of information available in the condition component on the diversity of solutions in CVAEs. This analysis helps to control the solution diversity depending on the problem requirements.\par We choose 3D object shape inverse rendering from single view as a complicated and ill-posed optimization problem. Reconstructing 3D volumes from 2D images, also known as \emph{3D inverse rendering}, is a challenging and ill-posed problem in computer vision with a nonlinear solution space. There may exist missing or hidden parts in the given 2D input images, which can be reconstructed in many 3D shapes. Effective regularizers are needed to solve such ill-posed problems, as a regularizer can apply appropriate constraints on the solution space to obtain promising results. \par For inverse rendering problem, we design a CVAE using the input 2D image as condition and analyze the effect of pose in 2D image, as the source of information, on the standard deviation of the output as a measure of computing diversity. In the experiments we evaluate our designed CVAE by comparing it to several recent related methods to show thet its promissing performance as a single view 3D reconstruction method and analyze the effect of condition on its prediction. \par

The remaining structure of this paper is organized as follows: Sec.~\ref{slr} includes a brief review of some recent related works about using CVAE for prediction and its analysis. Sec.~\ref{spd} starts with problem definition followed by our proposed CVAE structure for prediction. Sec.~\ref{se} contains experimental results to show the effect of the network condition on the prediction uncertainty in CVAE. Also evaluation of our designed CVAE is presented by comaring it to several recent metohds for single view 3D reconstruction. The paper is concluded in Sec.~\ref{sc}.
\section{Related Works}{\label{slr}}
In the recent deep learning literature, there exists a significant interest in using and analyzing CVAE for different optimization problems~\cite{deshpande2017learning, walker2016uncertain, greenwood2017predicting, vahdat2018dvae}. In this section we review some of the most recent studies on CVAE as the most related works to the subject under this study.\par
As some recent deep \textit{state-of-the-art} for 3D reconstruction, we name \cite{choy20163d}, in which a Recurent Neural Network is proposed for single an multiple view 3D reconstruction and \cite{wu2016learning, smith2017improved}, Generative Adversrial Networks are used for designing 3D reconstruction frameworks. The formar is a generative model and the latter is a single view 3D reconstruction method. \par
In~\cite{rezende2016unsupervised}, a CVAE-based framework is proposed with unsupervised training strategy for 3D object reconstruction, where the 3D shape is recovered from single or multiple views.  
In~\cite{sohn2015learning}, CVAE is used as a stochastic predictor for structural classification. The multi-modal distribution estimated by CVAE, which is due to its stochastic topology, leads to promising results in structural classification compared with similar deterministic predictors. The authors also reported some experiment to show that stochastic prediction in CVAE leads to high generalization power to the partial observations. \par

In~\cite{mattei2018leveraging}, Deep Latent Variable Models (DLVMs) are revisited to show that the  maximum-likelihood objective in these models is ill-posed for continuous problems and well-posed for discrete problems. Moreover, DLVMs can be related to non-parametric mixture models and take advantage of this potential to find an upper bound for CVAE objective. Finally,~\cite{mattei2018leveraging} proposes a method for handling missing data in CVAE prediction.\par

In~\cite{huang2018introvae} a new training technique for VAEs is proposed. This technique combines the strengths of GANs and VAEs and produces high-resolution photographic images. Besides,~\cite{shu2018amortized} indicates that \emph{amortized} inference in VAEs, (i.e. using empirical approximations instead of computing exact statistics) provides appropriate regularization for maximum likelihood estimate resulting in a good generalization. \par

To the best of our knowledge, controlling stochastic behaviour of CVAE has not yet been analyzed in the literature. Therefore, we propose to study and analyze the behaviour of CVAE in deterministic predictions.
\section{Problem definition}{\label{spd}}
In this section we first define CVAE as a regularized predictor using the concepts from~\cite{doersch2016tutorial} and then, demonstrate the designed CVAE structure for 3D shape inverse rendering.
\subsection{Using Conditional Variational Autoencoder (CVAE) as a predictor}
\label{scvae}
Formally, the main VAE objective is to maximize:
\begin{equation}
P(\underline{z})=\int P(\underline{z}|l;\theta)P(l)dl
\end{equation}
where, $P(\underline{z})$ is the data likelihood, $P(\underline{z}|l;\theta)$ is the output distribution of VAE which is often chosen to be Gaussian, i.e $P(\underline{z}|l;\theta)=N(\underline{z}|f(l;\theta) , \sigma^2*I)$ where, $f(z;\theta)$ stands for the VAE decoder output.\par
In order to avoid the regions where $l$ leads $P(\underline{z}|l;\theta)$ to become nearly zero, VAE introduces an encoder to control and restrict $P(l)$, by defining the distribution $Q(l|\underline{z};\beta)$ instead of $P(l)$, which uses the information from the target data to control the latent distribution. This policy in VAE increases the training speed and improves the feasibility of found solutions and is an appropriate mechanism to solve ill-posed optimization problems. \par
The output of VAE is $E_{l\sim Q}(P(\underline{z}|l;\theta))$, where the latent variable distribution, i.e., $Q(l|\underline{z};\beta)$, is constrained to be like the true probability $P(l|\underline{z})$, i.e., the probability of inferring $l$ form $\underline{z}$. This constraint is formulated using KL-divergence operator:
\begin{equation}
D(Q(l|\underline{z};\beta)||P(l|\underline{z}))=E_{l\sim Q}[logQ(l|\underline{z};\beta)-logP(l|\underline{z})]
\end{equation}
where $\beta$ and $\theta$ are parameters of trainable encoder and decoder, respectively.\par
Applying Bayes rule to $P(l|\underline{z})$ and changing term, we have:
\begin{equation}
\label{eq:kl}
\begin{split}
logP(\underline{z})-D(Q(l|\underline{z};\beta)||P(l|\underline{z})) = \\
 E_{l\sim Q}[logP(\underline{z}|l;\theta)]-D(Q(l|\underline{z};\beta)||P(l))]
\end{split}
\end{equation}

The left-hand side in (\ref{eq:kl}) is equivalent to the objective of VAE to be maximized and the right hand-side can be computed by VAE. In the VAE objective it can be observed that the KL-divergence term appears like a regularization term \cite{shu2018amortized}, which firstly, forces the distribution of the latent variable to be obtained from the target data and secondly, constrains the distribution to be Normal Gaussian. This KL-divergence term can be viewed as a generic regularizer in any optimization problem. Therefore, VAE can be considered as a regularized optimizer, for solving complicated optimization problems. It is worth noting that in KL-divergence  term, $P(z)$ is set to $N(0,I)$ assuming it can be converted to any distribution consistent with the latent variable in the trainable decoder \cite{b1}. \par
In the case of CVAE which is our focus in this paper, all terms in (\ref{eq:kl}) are conditioned on variable $\underline{m}$. Therefore the condition $\underline{m}$ will be fed into both decoder and encoder components. \par
Basically, CVAE is a stochastic optimizer because of the sampling component between encoder and decoder. Therefore it results some degree of uncertainty in the obtained solution. Our objective in this paper is to analyze the possibility of controlling the solution diversity by condition.
On other method of controlling the uncertainty in CVAE is the length of latent variable. No sampling results in a deterministic predictor and a high dimensional sampling results in appearing noise in prediction.  Since this variable includes prior information about the prediction, it will be useful to use it to obtain more accurate results. The optimum latent variable length can be found by validation methods. In this paper our focus is only on the condition as other resource for controlling the diversity of obtained solutions by fixing the dimensions of sampling component.\par
The network structure of our designed CVAE for prediction analysis will be illustrated in the following section.
\subsection{Network structure}
\begin{figure}[ht]

\centering
\begin{tabular}{c}
\includegraphics[scale = 0.35]{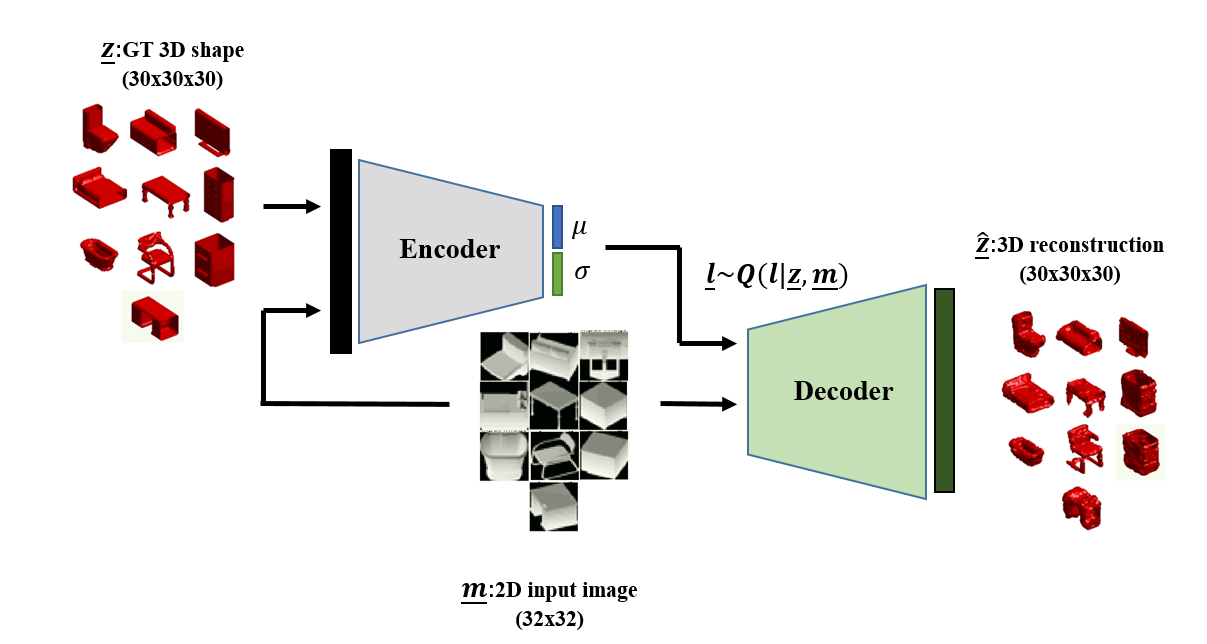}\\
(a) Train phase \\
\includegraphics[scale = 0.35]{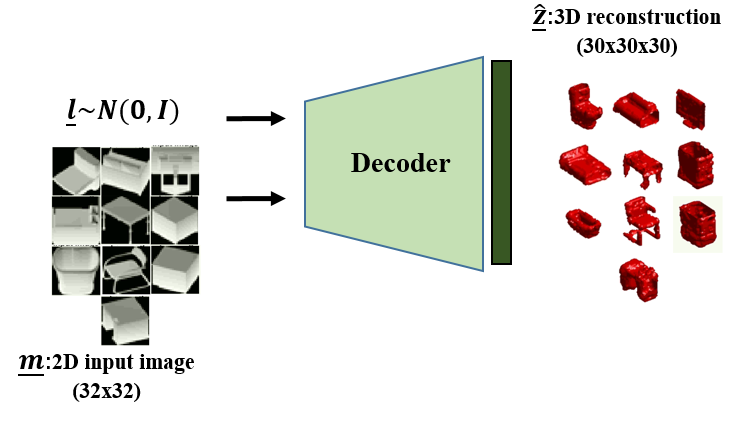}\\
(b) Test phase \\
\end{tabular}
\caption{CVAE structure for 3D shape inverse rendering. (a) An encoder is used to control latent distribution. (b) The encoder will be omitted in test phase.}
\label{figcvae}

\end{figure}

Based on the formulations in Section  \ref{scvae}, we design our CVAE structure for 3D  shape inverse rendering, shown in Figure \ref{figcvae}. 

In this structure which is also used in \cite{rezende2016unsupervised, sohn2015learning} for regression, the measurement, 2D image here,  is used as condition ($\underline{m}$) and prediction, 3D shape here, is used as $\underline{z}$.\par

\section{Experiments}{\label{se}}
In this section we evaluate our designed CVAE for 3D shape inverse rendering in two phases. In the first phase, we compare our designed CVAE with recent and \textit{state-of-the-art} methods for single view based 3D reconstruction methods from literature. Our main objective in this phase is to show the performance of our designed CVEA as an appropriate single view 3D reconstruction method comparable with exosting methods. In the second phase, 
we monitor the effect of informative condition on the solution diversity of CVAE for prediction. We introduce 2D input image as condition of CVAE and \textit{pose} as information in condition in this phase. \par
The following subsections include parameter setting and dataset used in our experiments followd by two designed phases.
\subsection{Parameter setting}
The detailed structure of encoder and decoder of proposed CVAE can be seen in Tables \ref{enc}, \ref{dec}, respectively. The structures are inspired by the structure used in \cite{wu2016learning} as a GAN for 3D reconstruction. We used keras for implementing our CVAE with 50 epochs and $ADAM$  optimizer with default learning-rate on a NVIDIA GeForce GTX 1080 Ti graphic processor. 
\begin{table}[h]
\caption{Details of Encoder Structure}
\label{enc}
\centering
\tiny{
\begin{tabular}{|c|c|}
\hline
Layer name &Shape\\
\hline
\textbf{\textit{Input}} & $32 \times 32 \times 36 \times 1$ \\
\hline
\textbf{\textit{Conv3D - padding = `same'}} & $3 \times 3 \times 3 \times 8$ \\
\hline
\textbf{\textit{Leakyrelu}} & $\alpha = 0.1$ \\
\hline
\textbf{\textit{Max pooling 3d}} &- \\
\hline
\textbf{\textit{Conv3D - padding = `same'}} & $3 \times 3 \times 3 \times 64$ \\
\hline
\textbf{\textit{Leakyrelu}} & $\alpha = 0.1$ \\
\hline
\textbf{\textit{Batch normalzation}} &- \\
\hline
\textbf{\textit{Max pooling 3d}} &- \\
\hline
\textbf{\textit{Conv3D - padding = `same'}} & $3 \times 3 \times 3 \times 128$ \\
\hline
\textbf{\textit{Leakyrelu}} & $\alpha = 0.1$ \\
\hline
\textbf{\textit{Batch normalzation}} &- \\
\hline
\textbf{\textit{Max pooling 3d}} &- \\
\hline
\textbf{\textit{Conv3D - padding = `same'}} & $3 \times 3 \times 3 \times 256$ \\
\hline
\textbf{\textit{Flatten}} &- \\
\hline
\textbf{\textit{Dense}} & 256 \\
\hline
\textbf{\textit{Leakyrelu}} & $\alpha = 0.1$ \\
\hline
\textbf{\textit{Batch normalzation}} &- \\
\hline
\textbf{\textit{Dropout}} & rate = 0.2 \\
\hline
\textbf{\textit{Dense}} & 128 \\
\hline
\textbf{\textit{Leakyrelu}} & $\alpha = 0.1$ \\
\hline
\textbf{\textit{Dense}} & 512 \\
\hline
\textbf{\textit{Dense ($\mu, \sigma$)}} & 32, 32 \\
\hline
\textbf{\textit{Lambda (random sampling)}} & 32 \\
\hline
\end{tabular}
}
\end{table}

\begin{table}[h]
\caption{Details of Decoder Structure}
\label{dec}
\centering
\tiny{
\begin{tabular}{|c|c|}
\hline
Layer name &Shape\\
\hline
\textbf{\textit{Dense}} & 256 \\
\hline
\textbf{\textit{Leakyrelu}} & $\alpha = 0.1$ \\
\hline
\textbf{\textit{Batch normalzation}} &- \\
\hline
\textbf{\textit{Dropout}} & rate = 0.2 \\
\hline
\textbf{\textit{Reshape}} & $4\times4\times4\times4$ \\
\hline
\textbf{\textit{Conv3D - padding = `same'}} & $3 \times 3 \times 3 \times 256$ \\
\hline
\textbf{\textit{Leakyrelu}} & $\alpha = 0.1$ \\
\hline
\textbf{\textit{Upsampling3D}} & $2\times2\times2$ \\
\hline
\textbf{\textit{Conv3D - padding = `same'}} & $3 \times 3 \times 3 \times 128$ \\
\hline
\textbf{\textit{Leakyrelu}} & $\alpha = 0.1$ \\
\hline
\textbf{\textit{Upsampling3D}} & $2\times2\times2$ \\
\hline
\textbf{\textit{Conv3D - padding = `same'}} & $3 \times 3 \times 3 \times 16$ \\
\hline
\textbf{\textit{Leakyrelu}} & $\alpha = 0.1$ \\
\hline
\textbf{\textit{Upsampling3D}} & $2\times2\times2$ \\
\hline
\textbf{\textit{Conv3D - padding = `same'}} & $3 \times 3 \times 3 \times 8$ \\
\hline
\textbf{\textit{Leakyrelu}} & $\alpha = 0.1$ \\
\hline
\textbf{\textit{Conv3D - padding = `same'}} & $3 \times 3 \times 3 \times1$ \\
\hline
\end{tabular}
}
\end{table}

\subsection{Dataset}

We used two popular 3D object datasets including Modelnet10 \cite{wu20153d}  and Shapenet  \cite{chang2015shapenet} in our experiments. In order to analyze the deterministic prediction of CVAE, we used Modelnet10 dataset and for the sake of comparison with other methods for single view based 3D reconstruction we used Shapenet dataset.\par
In the case of Modelnet10 dataset, we used the training and test sets defined by the dataset for our experiments.
In the case of Shapenet dataset, we used  4 classes including car, airplane, chair, couch for training and testing. The training and test sets for Shapenet are selected randomly. The size of input images to our designed CVAE is set to $128 \times 128 \times 4$ and we used the images with clean backgrounds for training and comparison.\par
\subsection{Perfromance comparison with \textit{state-of-the-art} and recent single view 3D reconstruction methods}
In this section, in order to have better evaluation of our designed CVAE for prediction, we compare it with several recent methods for single view 3D reconstruction including \cite {fan2017point}, called \emph{PSGenerator}, \cite{choy20163d}, called \emph{3DR2N2}, and \cite{groueix2018papier}, called \emph{AtlasNet}. For each method we used their pre-trained model available in the web without post-processing. Note that for \emph{3DR2N2}, we used the model made available by \cite{fan2017point}. In the case of the methods \emph{PSGenerator}, \emph{3DR2N2}, \emph{AtlasNet}, the comparisons are reported on the shapenet dataset for evaluation. The datasets are selected based on the evluation mechanism used in each compared method's paper.
\par
\begin{figure}
\hspace{-1cm}
\includegraphics[scale=0.4]{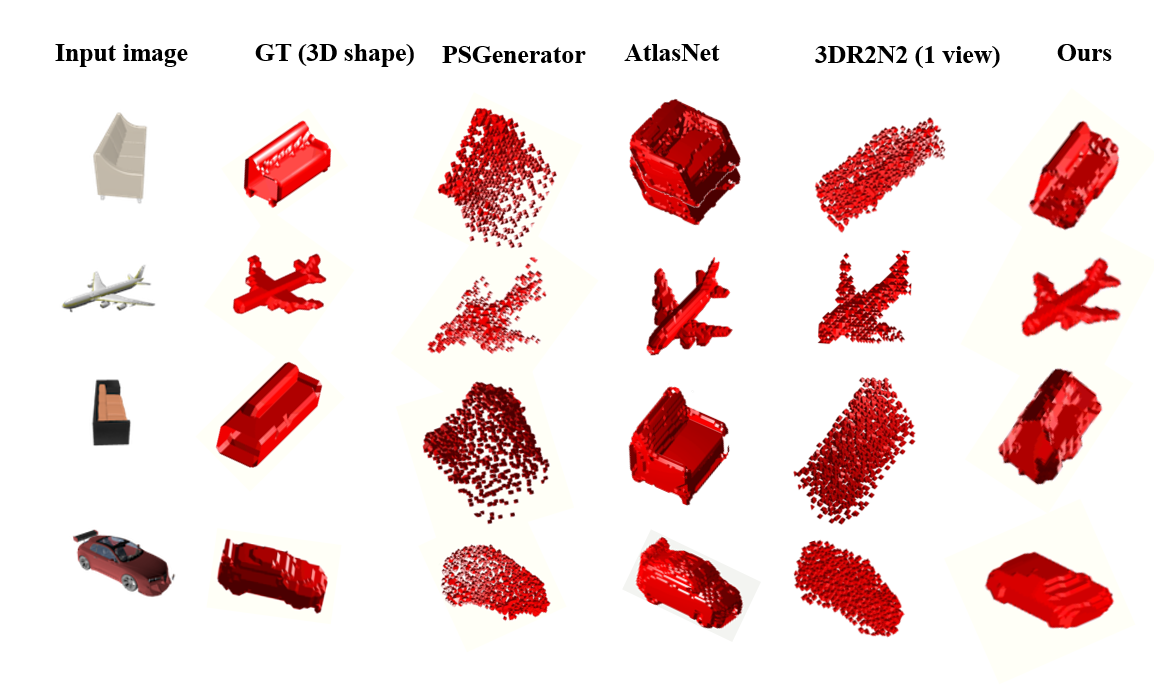}
\caption{Visual reconstruction results obtained by our designed CVAE and~\cite{fan2017point}. Four random test images are selected from different classes of shapenet dataset and are fed into networks. In the case of CVAE the output is computed using the mean of distribution of input noise to decoder.}
\label{f_cmp}
\end{figure}

 Figure \ref{f_cmp} shows the visual results of comparison between our designed CVAE to the work in \cite{fan2017point}, using the available network and weights from the web. Since the reconstruction of compared method are in the form of point clouds, we converted the output to $32 \times 32 \times 32$ voxels and then computed and showed the results.  \par

 As quantitative result, Table \ref{res} shows the numerical results obtained by designed CVAE and compared methods in terms of average shape intersection over union ($IOU$). Our results are averaged over 10 independent runs using 10 fixed predetermined equally spaced noise values as input to decoder in test phase. In the case of our CVAE, the numbers in parentheses denote the standard deviation of CVAE output for predetermined noise values.\par From Figure. \ref{f_cmp} and Table \ref{res}, it is observable that our proposed CVAE could achieve comparable results with \textit{state-of-the-art} and outperform them in several cases. Therefore It can be considered as an appropriate framework for single vie inverse rendering framework and it is a valid network for being analyzed as a 3D reconstruction tool. 

\begin{table}
\caption{Quantitative results in terms of shape $IOU$ obtained by our designed CVAE compared with the single view 3D reconstruction methods proposed in \cite{ fan2017point} trained on four classes of Shapenet dataset. The numbers in parenthesis are the standard deviation between 10 independent runs using 10 predetermined noise values as input to decoder in test phase.}
\label{res}
\hspace{-1cm}
\begin{tabular}{|c|c|c|c|c|c|}
\hline
& \multicolumn{4}{|c|}{\textbf{Average shape $IOU$}}&\\
\hline
\textbf{\textit{Method}} & \textbf{\textit{airplane}} & \textbf{\textit{car}}& \textbf{\textit{chair}}& \textbf{\textit{couch}} & \textbf{\textit{Mean}} \\
\hline
Ours&\tiny{ \begin{tabular}{c}0.5905 \\ ($\pm 0.1383$) \end{tabular}}&\tiny{\begin{tabular}{c}\textbf{0.8329} \\($\pm 0.0918$)\end{tabular}}&\tiny{\begin{tabular}{c}\textbf{0.5448}\\ ($\pm 0.1620$)\end{tabular}}&\tiny{\begin{tabular}{c}0.6904 \\($\pm 0.1385$)\end{tabular}}&\tiny{\begin{tabular}{c}0.6646\\ ($\pm 0.1326$)\end{tabular}}\\
\hline
3DR2N2&0.513&0.798&0.466&0.628&0.6012\\
\hline
AtlasNet & 0.5014&0.8201&0.4813&0.6911&0.6812\\
\hline
PSGenerator& \textbf{0.601}&0.831&0.544&\textbf{0.708}&\textbf{0.671}\\
\hline
\end{tabular}
\end{table}

\subsection{Analyzing the effect of condition on solution diversity}
\begin{figure}
\hspace{-1.7cm}
\begin{tabular}{c}
\includegraphics[scale =0.6]{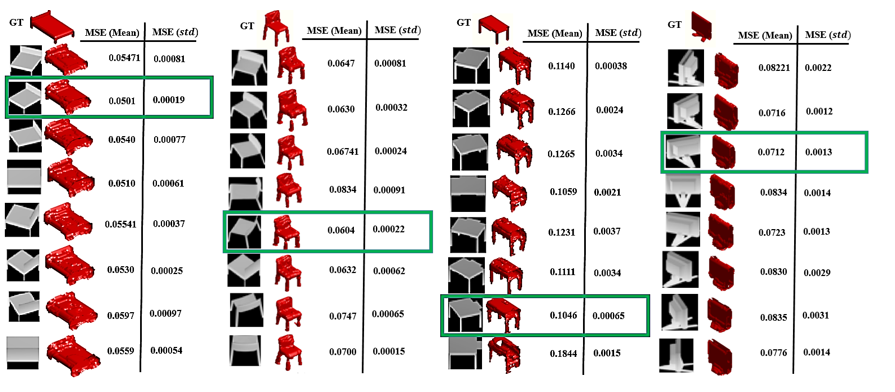}
\end{tabular}
\caption{CVAE average and standard deviation MSE results in 10 runs of four classes of Modelnet10 dataset for different poses. From left to right, bed, chair, desk and monitor classes are shown. The results with the least MSE are illustrated by green border as informative conditions.}
\label{figpose}
\end{figure}

After verifying the performance of our designed CVAE as a prediction framework, in this section the objective is to test the hypothesis \emph{"More informative condition results in lower solution diversity"}. By \emph{more informative condition}, we mean the condition containing more specific and adequate information about the prediction. For this aim we considered pose as information in 3D shape inverse rendering. Therefore, each object in training set of Modelnet10 dataset is rendered from 8 poses. Note that in order to omit the inter class overlap impact on the results, in this phase, for each class we trained a separate CVAE.
For instance, Figure~\ref{figpose} shows the reconstruction results obtained by CVAE trained for 4 classes separately. The results are averaged over 10 independent runs using 10 specified equally spaced noise values as the input to decoder in test phase Figure~\ref{figcvae}(b)).\par

From Figure~\ref{figpose}, it is observed that, for all object classes, considering the pose resulted in the least reconstruction error as informative pose, the most informative pose results in lower standard deviation. We should note that there sill exists intra-class overlap that affect the uncertainty of solutions.

\section{Conclusions and future works}{\label{sc}}

In this paper we focused on studying and analyzing the effect of information in condition on the diversity of prediction in CVAEs. We designed our CVAE for prediction by using the measurement as its condition component and used it for 3D object shape inverse rendering as a prediction problem. The experimental results show the promising performance of our designed CVAE compared with recent single view-based 3D reconstruction methods. By considering \emph{pose} as the information in the input images, the hypothesis \emph{``the more informative condition, the more deterministic the CVAE predictor''} is verified. 

This is an ongoing research and we plan to  analyze other elements that affect CVAE prediction, such as the number of training data and modality of the latent variable distribution.
\bibliography{references}
\bibliographystyle{plain}
\vspace{12pt}

\end{document}